\documentclass[journal]{IEEEtran}
\usepackage{cite}
\usepackage{amsmath,amssymb,amsfonts}
\usepackage{algorithmic}
\usepackage{graphicx}
\usepackage{dblfloatfix}
\usepackage{textcomp}

\graphicspath{{./figures/}}
\usepackage{multirow,booktabs,hyperref,url,bigstrut,pifont,microtype,subfigure}
\usepackage{xcolor}  
\newcommand{\lyj}{\textcolor{black}}
\newcommand{\ymj}{\textcolor{black}}
\newcommand{\lt}{\textcolor{black}}
\newcommand{\lm}{\textcolor{black}}

\newcommand{\revision}{\textcolor{black}}
\newcommand{\secondrevision}{\textcolor{black}}

\def\BibTeX{{\rm B\kern-.05em{\sc i\kern-.025em b}\kern-.08em
    T\kern-.1667em\lower.7ex\hbox{E}\kern-.125emX}}
\begin{document}
\title{EEGDM: Learning EEG Representation with Latent Diffusion Model}
\author{
Shaocong Wang$^{\dagger}$, 
Tong Liu$^{\dagger}$, 
Yihan Li, 
Ming Li, 
Kairui Wen, 
Pei Yang, 
Wenqi Ji, 
Minjing Yu, and Yong-Jin Liu$^{*}$
\thanks{This work was supported in part by the National Natural Science Foundation of China (U2336214, 62332019, 62461160309), and China Postdoctoral Science Foundation (2024M761678). ($^{*}$ Corresponding author: Yong-Jin Liu.)}
\thanks{Shaocong Wang, Tong Liu, Yihan Li, Ming Li, Kairui Wen, Pei Yang, Wenqi Ji and Yong-Jin Liu are with the MOE-Key Laboratory of Pervasive Computing, Department of Computer Science and Technology, Tsinghua University, Beijing 100084, China (e-mail: wangsc23@tsinghua.edu.cn; tongliu@tsinghua.edu.cn; liyihan23@mails.tsinghua.edu.cn; mingli\underline{~}thu@tsinghua.edu.cn; wkr21@mails.tsinghua.edu.cn; yangpei20@mails.tsinghua.edu.cn; jwq21@mails.tsinghua.edu.cn; liuyongjin@tsinghua.edu.cn). }
\thanks{Minjing Yu is with the College of Intelligence and Computing, Tianjin University, Tianjin 300350, China (e-mail: minjingyu@tju.edu.cn).}
\thanks{$^{\dagger}$  These authors contributed equally.}
}

\maketitle

\markboth{}{}

\begin{abstract}

Recent advances in self-supervised learning for EEG representation have largely relied on masked reconstruction, where models are trained to recover randomly masked signal segments. While effective at modeling local dependencies, 
\revision{the training objective of masked reconstruction does not compel the model to capture global generative constraints}
essential for characterizing neural activity. To address this limitation, we propose EEGDM, a novel self-supervised framework that leverages latent diffusion models to generate EEG signals as an objective. Unlike masked reconstruction, diffusion-based generation progressively denoises signals from noise to realism, compelling the model to capture holistic temporal patterns and cross-channel relationships. Specifically, EEGDM incorporates an EEG encoder that distills raw signals and their channel augmentations into a compact representation, 
\secondrevision{which serves as conditional information to guide the diffusion denoising process, thereby enabling the encoder and diffusion model to be jointly optimized through the generative objective.}
This design endows EEGDM with a compact latent space, which not only offers ample control over the generative process but also can be leveraged for downstream tasks. Experimental results show that EEGDM (1) reconstructs high-quality EEG signals, (2) learns robust representations, and (3) achieves competitive performance across diverse downstream tasks, 
thus exploring a new direction for self-supervised EEG representation learning.
\end{abstract}


\section{Introduction}
\label{Introduction}

\IEEEPARstart{E}{lectroencephalography} (EEG) is a non-invasive technique \lyj{that measures} spontaneous electrical activity \lyj{in} the brain \lyj{through scalp electrodes}. \lyj{The interpretation and utilization of EEG signals generally require sophisticated signal processing pipelines and advanced feature extraction methodologies} \cite{mohsenvand2020contrastive}. 
\lyj{Conventional EEG signal analysis typically employs manually engineered operations tailored to specific} EEG features of interest (e.g., band-pass filtering \lyj{in} a specific frequency range), \lyj{potentially excluding other informative signal components} \cite{zander2011towards}.

\lyj{Recently, deep learning has introduced a new research paradigm for EEG signal analysis due to its capability to automatically learn efficient representations from original data}
\cite{xu2022multichannel,wang2023motor,li2025deep, liu2026eeg}. 
\lyj{Current deep learning-based approaches, which predominantly rely on large-scale EEG datasets, can be broadly classified into two categories: (1) task-specific models, and (2) multi-task large models. Task-specific models face significant challenges in generalization due to substantial variations across datasets} in channel configurations, sampling rates, and \lyj{EEG recording} time lengths. \lyj{These discrepancies \ymj{necessitate} model training \ymj{on} individual datasets, thereby severely limiting cross-dataset generalizability. 
For multiple tasks, three large EEG models, EEGPT \cite{wangeegpt}, LaBraM \cite{jiang2024large}, and CBraMod \cite{wang2024cbramod} have recently been proposed. These models learn generic representations through masked reconstruction pre-training objective before being fine-tuned for various downstream tasks.}
\revision{However, the training objective of masked reconstruction does not compel the model to capture global generative constraints, as local interpolation from visible anchors often suffices -- particularly in EEG where strong local autocorrelation exists.}

\revision{In this paper, we propose EEGDM, a self-supervised framework that replaces the masked-reconstruction objective with diffusion-based EEG generation~\cite{ho2020denoising,dhariwal2021diffusion}. Unlike predicting masked segments -- a task solvable by local interpolation from visible anchors -- generating complete EEG signals from noise requires modeling the global data distribution. This generative objective compels the model to learn holistic temporal patterns and cross-channel relationships. Specifically, an EEG encoder transforms raw signals and their channel augmentations into compact latent representations. These representations condition a diffusion model to reconstruct the original EEG signals from noise, thereby endowing the encoder with rich and generalizable EEG semantics for downstream tasks.}
\secondrevision{By learning the encoder through a label-free conditional denoising objective, EEGDM encourages the representation to preserve structural information that is useful for modeling individual EEG samples, rather than optimizing representation quality according to downstream class frequency. This property may be particularly beneficial under severe class imbalance, as later evidenced by the class-balanced results on TUEV.}
	
\ymj{\lyj{We note that} EEG signals present unique challenges
due to their inherently low signal-to-noise ratio \lt{(SNR)} and complex characteristics, including stochasticity, nonstationarity, and nonlinearity \cite{gramfort2013time,cole2019cycle}. To address these challenges, we incorporate two \secondrevision{practical} components in our \lyj{EEGDM} framework. First, we employ channel augmentation that \lt{provides diverse training perspectives while enriching the conditional information.}
\revision{Second, we project EEG patches into a compact latent space via Principal Component Analysis (PCA) along the temporal dimension.}
\secondrevision{Interestingly, our ablation experiments show that PCA and EEG augmentation individually improve class-wise performance stability, while their joint use further improves overall class-balanced performance and representation discriminability. We therefore regard PCA and EEG augmentation as complementary training designs rather than standalone methodological contributions, and further analyze this behavior in Section~\ref{Ablation}.}
\lyj{Inspired by} the success of state-of-the-art \lt{DMs (e.g., Latent Diffusion Models (LDMs) \cite{rombach2022high} \lyj{and} Diffusion Transformers (DiTs) \cite{peebles2023scalable}) operating in latent spaces, we apply our DM to}
this PCA-derived latent space rather than the original \secondrevision{high-dimensional} EEG space. 
This enables EEGDM to focus on learning meaningful signal patterns for effective representation learning across diverse downstream tasks. \lyj{Our} experiments \lyj{show} that EEGDM effectively learns robust representations, achieving competitive performance across diverse downstream tasks 
and exhibiting strong generalizability.}

Our contributions are summarized as follows:
\begin{itemize}
	\item \revision{We propose EEGDM, a self-supervised framework that leverages diffusion-based EEG signal generation as a representation learning objective. By synthesizing signals from noise, EEGDM fundamentally differs from masked-reconstruction approaches and learns representations that capture holistic EEG semantics.}
	\item \secondrevision{We design a jointly optimized encoder–diffusion architecture in which the EEG representation produced by the encoder serves as conditional information for the latent diffusion model. By back-propagating the denoising objective through this conditional representation, the encoder and diffusion model are jointly trained to learn inherent EEG representations.}
	\item \revision{We validate the effectiveness and generalizability of EEGDM across six diverse EEG analysis tasks.} 
    \secondrevision{Notably, EEGDM exhibits strong robustness to severe class imbalance, achieving substantially improved class-balanced metrics and minority-class recall on TUEV. In addition, its diffusion module enables synthetic EEG generation for downstream data augmentation, as validated on CL-Drive.}
\end{itemize}

The remainder of this work is organized as follows. First, we describe the related works in Section \ref{section2}. Next, the proposed algorithm is introduced in detail in Section \ref{section3}. After that, experimental results are shown in Section \ref{section4}. Finally, the conclusion is given in Section \ref{section5}.

\section{Related work}
\label{section2}
\subsection{Self-supervised EEG Representation Learning}

\lyj{Self-supervised learning can leverage large amounts of unlabeled data for pre-training, thereby reducing the reliance on manual annotations. This methodology has led to significant breakthroughs in natural language processing and computer vision \cite{kenton2019bert,he2022masked}. However, its full potential in EEG analysis remains largely unexplored \cite{jiang2024neurolm}.}

Some \lyj{studies} \cite{mohsenvand2020contrastive,banville2021uncovering} \lyj{have explored} learning representations of EEG signals via contrastive learning, which trains a channel feature extractor by extending the SimCLR framework \cite{chen2020simple} to time series data. Another \lyj{line} of work extends the mask idea to EEG analysis \cite{mohammadi2024eeg2rep}. For \lyj{instance}, 
Chien et al. \cite{chien2022maeeg} propose a masked auto-encoder, which learns the EEG representation by learning to reconstruct the masked EEG features using a Transformer architecture. 
BENDR \cite{kostas2021bendr} uses a convolutional encoder to extract EEG features from local time windows, masks \lyj{a portion of them}, and \lyj{utilizes a Transformer to predict} the information in the masked part. 
\lt{ LaBraM \cite{jiang2024large} pre-trains \lyj{a} Transformer \lyj{model by predicting} the original neural codes of masked EEG channel patches \lyj{based on} a defined neural codebook. EEGPT \cite{wangeegpt} introduces \lyj{the alignment of} spatio-temporal representation, implemented by a mask-based dual self-supervised learning for efficient extraction \lyj{of EEG features}.}
CBraMod \cite{wang2024cbramod} introduces a structurally aware EEG foundation model that separately captures heterogeneous spatial–temporal dependencies through a criss-cross transformer and achieves strong cross-dataset generalization via large-scale masked pre-training.


Unlike the above approaches, we explore 
self-supervised EEG representation learning via generative models. \lyj{Our} proposed EEGDM not only \lyj{facilitates high-quality} EEG generation but also \lyj{enables the} learning \lyj{of} rich and robust representations \lyj{for downstream tasks}.

\subsection{Diffusion Model}
\ymj{Diffusion models learn complex data distributions through a progressive denoising process that requires a deep understanding of the underlying data structure and semantics. This fundamental characteristic endows them with dual capabilities: 
	\lt{generating} high-quality samples and \lt{learning} meaningful representations \lyj{inherent in the data distribution}. 
	The \lyj{pioneering} work by Sohl-Dickstein et al. \cite{sohl2015deep} and subsequent advances by Ho et al. \cite{ho2020denoising} established diffusion models as powerful tools, leading to widespread applications in generative tasks \cite{rombach2022high,peebles2023scalable,zhang2023adding}. More recently, \lt{their potential for representation learning has gained attention, with emerging studies}
	showing promising results in learning discriminative features \cite{mittal2023diffusion,chen2024deconstructing,hudson2024soda}.}

In the EEG domain, existing diffusion-based approaches have \lyj{primarily} focused on generative applications, such as EEG synthesis and data augmentation \cite{huang2024eegdfus,chen2024improving,huang2025sad}, or \lt{a few} specific supervised tasks \cite{kim2023diff,kim2024brain}.
\revision{Recently, Puah et al. \cite{puah2025eegdm} proposed an EEG representation learning framework, also termed EEGDM. To avoid ambiguity, we refer to their method as {\it Puah-EEGDM-SSM}. Their approach integrates a Structured State-Space Model (SSM) into the diffusion pre-training paradigm to better capture the temporal dynamics of EEG signals. In contrast, our EEGDM adopts a fundamentally different architectural paradigm. Specifically, we employ a Latent Diffusion Transformer (DiT) operating on a refined latent manifold, which is specifically designed to capture global inter-channel spatial synchrony and complex phase relationships—critical neural features that are overlooked by the linear recurrence mechanisms inherent in SSM.}

\begin{figure*}[ht]
	\begin{center}
		\centerline{\includegraphics[scale=0.70]{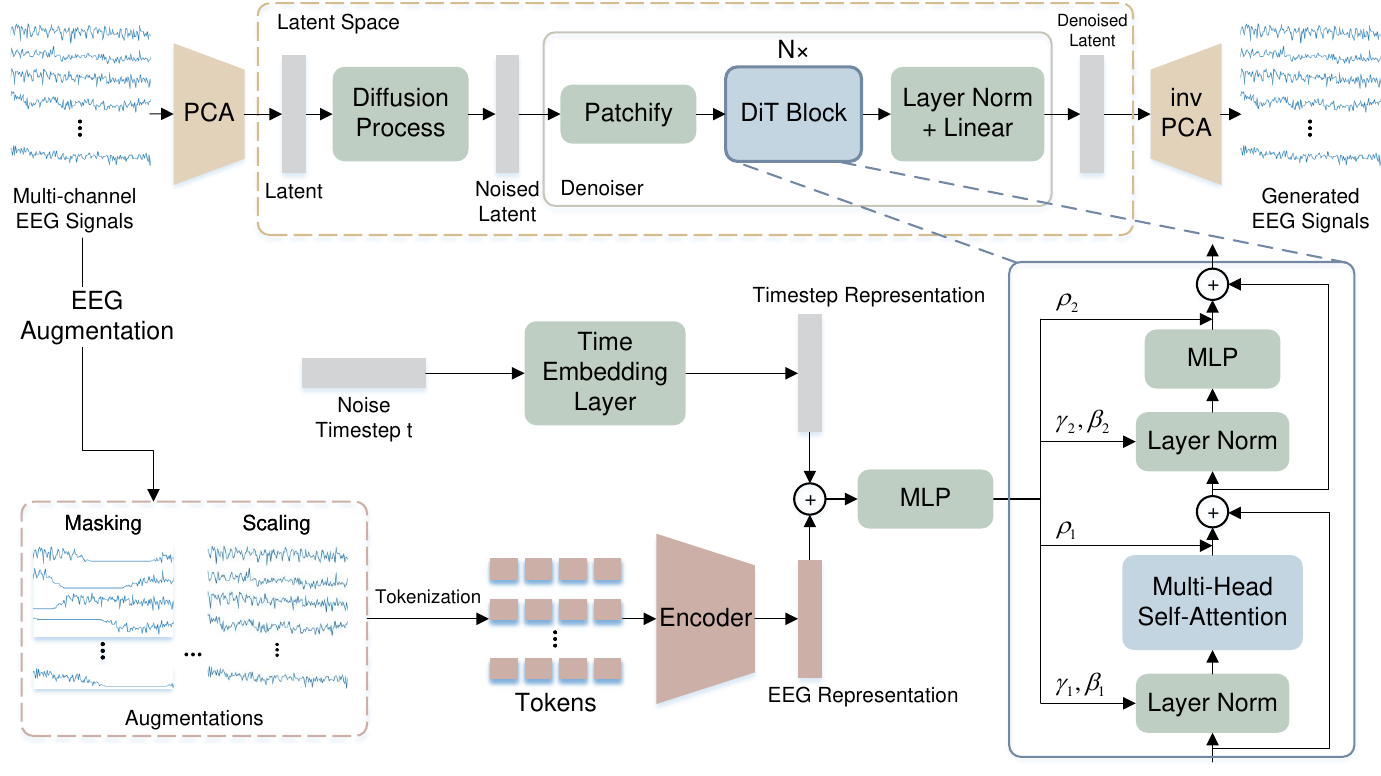}}
        \caption{The overall architecture of EEGDM. The original EEG data is projected onto a latent space using PCA. Noise is added in this latent space, and the denoised latent representation is predicted by learning the conditional latent diffusion model. The prediction is then projected back to the original space by inverse PCA to obtain the generated target EEG signal. The encoder transforms the source EEG signal (including original data and augmentations) into a concise EEG representation. It is combined with the noise timestep representation and then used as conditional information to modulate the DiT block, thus guiding the denoising of the latent representation. \revision{In the adaLN-Zero blocks, $\rho_1,\rho_2$ denote the dimension-wise scaling parameters applied before residual connections, and $\gamma_1,\gamma_2,\beta_1,\beta_2$ denote the adaLN scale and shift parameters, respectively.}}

		\label{Figure 1}
	\end{center}
\end{figure*}

\section{Method}
\label{section3}
\lyj{Our study is motivated by the following observation: Diffusion models capable of generating rich and high-fidelity content from scratch are likely to effectively characterize the underlying properties, structures, and dynamics of the data during the generative process.} 
Therefore, we explore self-supervised EEG representation learning via the diffusion model, \lyj{and design the following framework architecture.}

\revision{For clarity, Table \ref{tab:symbols} lists the principal symbols used in the following sections.}

\begin{table}[h]
\centering
\caption{Symbol table.}
\label{tab:symbols}
\begin{tabular}{ccl}
\toprule
Symbol & Domain & Definition \\
\midrule
$x$ & $\mathbb{R}^{C\times t_s}$ & Multi-channel EEG sample \\
$x'$ & $\mathbb{R}^{C\times t_s}$ & Source EEG signal (original + augmentations) \\
$z$ & $\mathbb{R}^{K}$ & PCA-compressed latent representation \\
$z_t$ & $\mathbb{R}^{K}$ & Noisy latent at timestep $t$ \\
$\epsilon_t$ & $\mathbb{R}^{K}$ & Sampled Gaussian noise \\
$\epsilon_\theta$ & -- & Denoiser (noise prediction network) \\
$e$ & $\mathbb{R}^{d}$ & Aggregated conditional EEG representation \\
$t$ & $\mathbb{N}$ & Diffusion timestep \\
$\alpha_t$ & $\mathbb{R}$ & Noise schedule coefficient \\
$\bar{\alpha}_t$ & $\mathbb{R}$ & Cumulative product of $\alpha_s$ \\
$\theta$ & -- & Parameters of the diffusion model \\
$\rho_1,\rho_2$ & $\mathbb{R}^{d}$ & adaLN-Zero dimension-wise scaling \\
$\gamma_1,\gamma_2$ & $\mathbb{R}^{d}$ & adaLN scale parameters \\
$\beta_1,\beta_2$ & $\mathbb{R}^{d}$ & adaLN shift parameters \\
$K$ & $\mathbb{N}$ & Number of PCA components ($K=20$) \\
$C$ & $\mathbb{N}$ & Number of channels \\
$t_s$ & $\mathbb{N}$ & Length of each EEG segment (time points) \\
$\omega$ & $\mathbb{N}$ & Patch length (time points) \\
$d$ & $\mathbb{N}$ & Token embedding dimension ($d=512$) \\
$N$ & $\mathbb{N}$ & Number of DiT blocks \\
$\phi$ & -- & Parameters of the EEG encoder $f_\phi$ \\
$\mu_\theta$ & -- & Predicted reverse-process mean \\
$\Sigma_\theta$ & -- & Predicted reverse-process covariance \\
$\lambda_{\mathrm{vlb}}$ & $\mathbb{R}$ & Weight of the variational lower-bound loss ($=1$) \\
$T$ & $\mathbb{N}$ & Total number of diffusion timesteps ($T=1000$) \\
\bottomrule
\end{tabular}
\end{table}

\subsection {Framework Architecture}
\lyj{Our} proposed self-supervised EEG representation learning \lyj{framework}, EEGDM, is \lyj{formulated as} a latent diffusion model (DM) conditioned on EEG \secondrevision{representations extracted from original and augmented source views.}
Our goal is to generate high-quality EEG signals while \lyj{equipping the model with a generalizable understanding of EEG data that can be effectively adapted} to various downstream tasks. 
\secondrevision{Directly performing diffusion in the original high-dimensional EEG space is challenging. We therefore project EEG patches into a compact PCA subspace and perform the denoising process in this latent target space. Importantly, PCA and EEG augmentation act on different branches of EEGDM: PCA transforms the denoising target, whereas the augmentations are used to construct the source views provided to the EEG encoder.}
\lyj{The overall architecture of EEGDM is illustrated in Fig. \ref{Figure 1}.}

Specifically, given the original multi-channel EEG signals $X\in\mathbb{R}^{C\times T}$, where $C$ is the number of EEG electrodes (channels) and $T$ is the total timestamp, we segment $X$ into ${\lfloor{\frac{T-t^s}{s^t}}\rfloor}+1$ samples. This segmentation assumes a timestamp length\footnote{\lyj{Here superscript is used; subscript is reserved for diffusion formation in subsequent sections.}} $t^s$ per sample and a stride $s^t$. As shown in Fig. \ref{Figure 1},  each multi-channel sample $x\in\mathbb{R}^{C\times{t^s}}$ is processed by PCA \revision{along the temporal dimension} to obtain the latent representation $z$. \lyj{Our} diffusion model operates on this latent space, where noise is added and predicted. The denoised latent representation predicted by the diffusion model is projected back to the original space by inverse PCA to obtain the target EEG signal. Owing to the effectiveness of the DiT model in the latent space, we utilize it to predict the denoised latent representations. 


Furthermore, we introduce an EEG encoder \secondrevision{$f_\phi$} that transforms the input source signal $x^{\prime}$ (including original EEG sample and its augmentations) into a compact EEG representation. \secondrevision{This representation, together with the noise timestep embedding, serves as the conditional information for the diffusion model and modulates the DiT blocks during denoising. Specifically, for a source view $x'$, we have
\[
e_i=f_\phi(x'_i), \qquad
\hat{\epsilon}_t=\epsilon_\theta(z_t,t,e),
\]
where $\phi$ and $\theta$ denote the parameters of the EEG encoder and diffusion denoiser, respectively, and $e$ denotes the aggregated conditional representation constructed from the source-view representations $\{e_i\}$, as detailed in Section~\ref{Conditional Guidance}.
Importantly, the conditional representation $e$ is not detached during pre-training. Therefore, the denoising objective is back-propagated through $e$, jointly optimizing $\phi$ and $\theta$. In this way, representation learning in the encoder is directly coupled with the latent denoising objective.}

\begin{figure*}[h]
	\vskip 0.2in
	\begin{center}
		\centerline{\includegraphics[scale=0.75]{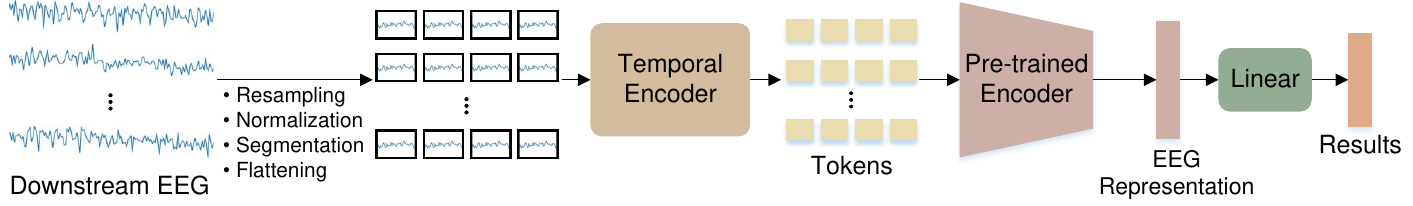}}
		\caption{The pre-trained encoder is connected to a linear prediction head to perform downstream tasks.}
		\label{Figure 2}
	\end{center}
\end{figure*}

\subsection{The EEG Encoder}

The representation output from the EEG encoder lies at the core of EEGDM, which (1) acts as the communication bridge between the encoder and the denoiser $\epsilon_\theta$, and (2) \secondrevision{provides informative conditional cues for the denoising process. Specifically, the encoder representation is extracted from a clean or augmented source EEG signal that is semantically related to the denoising target. It guides the diffusion model to predict the injected noise and recover the target latent representation from noisy inputs.}

\lyj{The core motivation behind the above EEG encoder design arises from the inherently under-determined nature of the denoising task. The denoiser tends to exploit any relevant knowledge or informative cues to infer and recover the original data distribution as accurately as possible. This incentivizes the encoder to distill salient commonalities between the source and target EEG signals into a compact representation. We constrain the encoded EEG representation to a low-dimensional latent space and guide the denoiser through feature modulation. This strategy encourages the encoder to capture high-level semantic information in the EEG signals, while leaving the reconstruction of local details to the denoiser. Rather than reconstruction from scratch as most conventional denoisers do \cite{hinton2006reducing,higgins2017beta}, our strategy enables the encoder to learn a representation that supports latent space denoising. This alleviates the burden of compressing all signal information into the representation, allowing it to prioritize the most distinctive and informative features of the EEG. Such a representation is particularly valuable for adapting to a wide range of downstream EEG tasks.}

Our EEG encoder is based on the Vision Transformer (ViT) architecture \cite{dosovitskiy2020image} that operates on the sequences of tokens. We segment each EEG channel into tokens and then flatten the tokens to form consistent biosignal sequences. We retain many of ViT's best practices, enabling the EEG encoder to be a Transformer architecture that can process input EEG signals with various formats (different sampling rates, number of channels, and time lengths). The input of the encoder is the source EEG signal $x^{\prime}$, which consists of the original samples and the EEG channel augmentations. EEG augmentations are a set of transformations that preserve the semantic information in the EEG channels without altering the interpretation of the EEG data. Here, we employ two augmentations (zero-masking and amplitude scaling) \lyj{which} were shown in \lyj{prior} work \cite{mohsenvand2020contrastive} to be the most effective \lyj{for} extracting useful features. \secondrevision{Specifically, zero-masking sets a contiguous 0.4-second temporal segment in each channel to zero, while amplitude scaling multiplies the entire signal by a factor sampled uniformly from $[0.8,1.2]$. For each EEG sample, we generate two augmented views during pre-training: one using zero-masking and the other using amplitude scaling. Together with the original EEG sample, these form three source views for conditional representation learning.}

We then use a $\omega$-length window without overlap to segment each EEG channel sample into patches ($x^{\prime}=\{x^{\prime}_{j,k}\in\mathbb{R}^{\omega}|j=1,2,\cdots, C,k=1,2,\cdots,{\lfloor{\frac{t^s}{\omega}}\rfloor}\}$). The total number of patches for $x^{\prime}$ is ${\left|x^{\prime}\right|}=C{\lfloor{\frac{t^s}{\omega}}\rfloor}$. Then a temporal encoder is used to encode each EEG patch into a patch embedding $e^{\prime}=\{e^{\prime}_{j,k}\in\mathbb{R}^{d}|j=1,2,\cdots, C,k=1,2,\cdots,{\lfloor{\frac{t^s}{\omega}}\rfloor}\}$, where $d$ is the dimension of the embeddings. 
\secondrevision{Specifically, the temporal encoder consists of a single one-dimensional convolutional layer with 512 output channels, a kernel size of 200, a stride of 200, and no padding, mapping each single-channel EEG patch to a $d=512$-dimensional token embedding. A learnable positional embedding is then added to each patch embedding, and the resulting token sequence is fed into an 8-block Transformer encoder with a hidden dimension of 512, 8 attention heads per block, and an MLP hidden dimension of 2048.}
We use average pooling on the output embeddings to obtain the EEG representation, which is then combined with the noise timestep representation to guide the denoising of the diffusion model. In addition, we perform downstream EEG tasks by using task-specific prediction heads on the EEG representation. As shown in Fig. \ref{Figure 2}, we train a linear predictor as the prediction head based on the pre-trained encoder in fine-tuning and then analyze the model's performance \lyj{on} the downstream tasks, \lyj{in which} we employ task-specific loss functions for different downstream objectives. The implementation details can be found in the subsection \ref{Experimental_Setup}.

\subsection{EEG Data Generation by Diffusion Model}

As aforementioned, \secondrevision{directly modeling the original temporal EEG space results in a high-dimensional denoising target. Following the LDMs \cite{rombach2022high}, we therefore perform diffusion in a compact PCA subspace. The PCA projection retains the dominant variations of each EEG patch while reducing the dimensionality of the target seen by the diffusion model. In addition, the representation produced by the EEG encoder is used as conditional information to guide the diffusion process.} After denoising in the latent space, the predicted latent representation is projected back to the original signal space using the PCA basis to reconstruct the final EEG signal.

\subsubsection{Diffusion Formulation}
The input \lyj{to} our proposed latent diffusion model, \lyj{which is} conditioned on channel augmentation, is the latent representation $z=Px$ \lyj{obtained through} PCA pre-processing. Here, we use the $\omega$-length window to segment each $x$ before applying PCA to \lyj{obtain the corresponding latent representation}. The PCA \lyj{basis vectors are} computed \lyj{via} eigen-decomposition \lyj{without the} need for gradient-based training. Our diffusion model operates \lyj{within} this latent space. The noisy input $z_t$ at a given timestep is transformed into patches, which are then linearly embedded into a sequence of tokens with dimension $d$ (implemented by the patchify layer).
Here, we apply a sine-cosine version of the positional embeddings to all input tokens. Then, the input tokens are processed by several DiT blocks. By applying the reparameterization trick, we can \lyj{do the sampling as follows}:
\begin{equation}
	\begin{aligned}	z_t=\sqrt{\bar{\alpha}_t}z_0+\sqrt{1-\bar{\alpha}_t}\epsilon_t,
	\end{aligned} \label{eq2}
\end{equation}
where $\epsilon_t\sim{\mathcal{N}(0,\mathbf{I})}$. \secondrevision{Given the conditional EEG representation $e$ produced by the EEG encoder,} the diffusion model is trained to learn the reverse \lyj{denoising} process:
\begin{equation}
    p_\theta(z_{t-1}\mid z_t,e)
    =
    \mathcal{N}\!\left(
    \mu_\theta(z_t,t,e),
    \Sigma_\theta(z_t,t,e)
    \right),
\label{eq3}
\end{equation}
\secondrevision{where $e\in\mathbb{R}^{d}$ denotes the EEG representation used as conditional information for the denoiser. The network predicts the mean $\mu_{\theta}$ and diagonal covariance $\Sigma_{\theta}$ of the conditional reverse distribution based on the noisy latent $z_t$, diffusion timestep $t$, and EEG condition $e$.}

\secondrevision{Following the noise-prediction parameterization, the denoiser $\epsilon_{\theta}$ predicts the injected noise conditioned on $z_t$, $t$, and $e$. It is optimized using the mean squared error between the predicted noise and the sampled Gaussian noise $\epsilon_t$:}
\begin{equation}
\mathcal{L}_{\mathrm{simple}}(\theta,\phi)
=
\mathbb{E}_{x,t,\epsilon_t}
\left[
\left\|
\epsilon_{\theta}(z_t,t,e)-\epsilon_t
\right\|_2^2
\right],
\label{eq4}
\end{equation}
\secondrevision{where $\phi$ denotes the parameters of the EEG encoder. Since the conditional representation $e$ is produced by the encoder and is not detached during pre-training, $\mathcal{L}_{\mathrm{simple}}$ is back-propagated through $e$, thereby jointly optimizing the encoder parameters $\phi$ and the diffusion-model parameters $\theta$.}

\secondrevision{To learn the reverse-process distribution with a predicted covariance, we additionally employ the variational lower-bound objective:
\begin{equation}
    \begin{aligned}
        \mathcal{L}_{\mathrm{vlb}}
        = \mathbb{E}_{q}\Bigg[
        &-\log p_{\theta}(z_0|z_1,e) \\
        &+\sum_{t=2}^{T}
        D_{\mathrm{KL}}\Big(
        q(z_{t-1}|z_t,z_0)
        \,\|\, 
        p_{\theta}(z_{t-1}|z_t,e)
        \Big)
        \Bigg].
    \end{aligned}
\label{eq:vlb}
\end{equation}
where $q(z_{t-1}|z_t,z_0)$ denotes the posterior induced by the predefined forward diffusion process and $D_{\mathrm{KL}}$ denotes the Kullback--Leibler divergence. Following the approach of Nichol et al. \cite{nichol2021improved}, the overall pre-training objective is
\begin{equation}
\mathcal{L}
=
\mathcal{L}_{\mathrm{simple}}
+
\lambda_{\mathrm{vlb}}\mathcal{L}_{\mathrm{vlb}},
\qquad
\lambda_{\mathrm{vlb}}=1.
\end{equation}
}


After the final DiT block, we use a standard linear decoder to decode the token sequence into an output noise prediction and an output diagonal covariance prediction. \lyj{Both outputs have the same data shape as the input}. 
Once $p_\theta$ is trained, sampling starts from
$z_T\sim\mathcal{N}(0,I)$ and iteratively samples
$z_{t-1}\sim p_\theta(z_{t-1}\mid z_t,e)$ for $t=T,\ldots,1$.
Finally, we obtain the denoised latent representation $\widetilde{z}$, \lyj{which is then} projected back to the original \lyj{signal} space \lyj{using} inverse PCA to \lyj{reconstruct} the original EEG signals:
\begin{equation}
	\begin{aligned}
		\widetilde{x}=P^{-1}\widetilde{z}=P^T\widetilde{z}.
	\end{aligned} \label{eq5}
\end{equation}

\subsubsection{Conditional Guidance}
\label{Conditional Guidance}
\lm{We incorporate the encoded EEG representation as conditional input for the diffusion model. It guides the generative process and is further leveraged for downstream tasks.}
\secondrevision{To enrich the conditional information, we construct multiple semantically consistent source views through EEG augmentation, following the multi-view learning strategy commonly used in contrastive learning~\cite{chen2020simple}.
Specifically, let $\{x'_i\}_{i=1}^{m}$ denote the source views of an EEG
sample. Each view is processed by the shared EEG encoder $f_\phi$:
\[
e_i = f_\phi(x'_i), \qquad i=1,\ldots,m.
\]
The resulting representations are aggregated by average pooling to form
the conditional EEG representation:
\[
e=\frac{1}{m}\sum_{i=1}^{m} e_i.
\]
In our implementation, $m=3$, including the original signal and two
augmented views.}

\secondrevision{The aggregated representation $e$ is then used as the conditional input of the reverse denoising process defined in Eq.~\ref{eq3}. During sampling, the denoiser directly predicts the injected noise conditioned on $z_t$, $t$, and $e$, such that the encoded EEG representation continuously guides the reverse diffusion process.}


\subsubsection{Conditional Input Setup}

\lyj{The treatment of conditional inputs and the initialization process significantly influence the generation quality in diffusion models.} We adopt the adaLN-Zero block from DiT to incorporate the EEG representation $e$ and diffusion timestep information into the denoiser. \secondrevision{The denoiser consists of 8 DiT blocks with a hidden dimension of 512 and 8 attention heads per block.}

\secondrevision{Specifically, the EEG representation $e$ is combined with the timestep embedding, and the resulting conditional embedding is used to regress the scale and shift parameters $\gamma$ and $\beta$ in the Transformer blocks.}
Additionally, we regress the dimension-wise scaling parameters \revision{$\rho$}, which are applied just before each residual connection in the DiT blocks. The MLP for all \revision{$\rho$} is initialized to output the zero vector, \lyj{which effectively initializes} the entire DiT block as an identity function. 
In addition to layer normalization, we incorporate conditional embeddings into residual connections, \lyj{which} enhance the vanilla adaLN and improve the quality of the generated EEG signals.

\section{Experiment}
\label{section4}

\subsection{Datasets}
\label{Datasets}

\textit{Pre-training datasets.} We utilize the Temple University Hospital EEG Corpus (\textbf{TUEG}) \cite{obeid2016temple} to pre-train the proposed EEGDM. TUEG is a very large publicly available clinical EEG dataset containing clinical-grade recordings collected in real-world hospital environments. It has over 40 different channel configurations and varying temporal durations of recordings. The majority of the recordings are acquired at a sampling rate of 256 Hz. TUEG exhibits substantial heterogeneity in patient demographics, channel configurations, and recording equipment, making it particularly suitable for pre-training our model.

\revision{\textit{Downstream tasks and Datasets.}
To comprehensively evaluate our method, we consider multiple downstream tasks, including event type classification (TUEV dataset \cite{harati2015improved}), abnormal detection (TUAB dataset \cite{lopez2015automated}), motor imagery classification (SHU-MI dataset \cite{ma2022large}), cognitive load assessment (CL-Drive dataset \cite{angkan2024multimodal}), emotion recognition (Seed-VII dataset \cite{jiang2024seed}), and emotion distribution prediction (DMER dataset \cite{yang2024multimodal}). Detailed descriptions of these datasets are provided in the supplementary material.}

It is worth noting that we primarily select downstream datasets from the past five years, a strategic shift from the benchmarks used in works like CBraMod \cite{wang2024cbramod} or EEGPT \cite{wangeegpt}. This decision stems from the need to evaluate the model against the fidelity and complexity of modern neuroscience standards. Older datasets often lack the high spatial density and signal quality required to stress-test a foundation model, whereas recent paradigms — ranging from naturalistic stimuli to multimodal retrieval — better assess the capacity for generalizable, high-level cognitive decoding. However, to ensure fair comparison with established baselines, we make exceptions for the TUAB and TUEV datasets. Despite falling outside the five-year window, these two widely adopted benchmarks are retained as standardized anchors, allowing us to validate our model’s performance on classical clinical tasks alongside its capabilities on modern, complex scenarios.

\subsection{Experimental Setup}
\label{Experimental_Setup}
For the pre-training dataset TUEG, following prior work \cite{wang2024cbramod}, we select 19 common EEG channels. All EEG signals are resampled to 200 Hz and segmented into non-overlapping 30-second samples (19 channels × 6,000 time points). Each EEG patch is set to a temporal length of 1 second, resulting in 570 patches per sample. For all downstream datasets, we maintain consistency with our pre-training data by resampling EEG signals to 200 Hz and setting each EEG patch duration to 1 second.

For the downstream datasets TUEV and TUAB, consistent with previous studies \cite{wangeegpt, jiang2024large, wang2024cbramod}, we utilize 16 common bipolar montage channels from the international 10-20 system, with sample durations of 5 seconds and 10 seconds, respectively. We adopt the train-test splits provided by these datasets. For the CL-Drive dataset, we use the channel configuration and sample duration (10 seconds) provided by the dataset and apply 10-fold subject-wise cross-validation. For the SHU-MI, DMER, and SEED-VII datasets, we use the channel configurations provided by each dataset and segment the signals into 4-second samples. We employ a subject-independent split strategy for these datasets. We fine-tune EEGPT \cite{wangeegpt}, LaBraM-Base \cite{jiang2024large} (the only variant with available pre-trained weights), and CBraMod \cite{wang2024cbramod} based on their publicly available code and pre-trained weights, except when their experimental results are already reported in the original papers.

For the diffusion module and encoder, we adopt the DiT and ViT architectures, respectively. \revision{All parameters in EEGDM are randomly initialized at the start of pre-training; no pre-trained weights from other models are used. The encoder and diffusion denoiser are jointly trained from scratch on TUEG.} Following common practices in generative modeling literature, we maintain an exponential moving average (EMA) of model weights with a decay rate of 0.9999 during pre-training. The specific hyperparameters for pre-training are detailed in the Supplementary Material (Table I). \revision{Pre-training is conducted on 4 $\times$ NVIDIA A100 GPUs with a batch size of 32. Training converges after approximately 329,400 steps (about 113 wall-clock hours).}

\begin{table*}[ht]
\centering
\caption{Performance comparison on TUEV. The best metric is highlighted in bold, and the second-best metric is underlined.}
\begin{tabular}{lccccccc}
\toprule
Methods & Params & Balanced Accuracy $\uparrow$ & Weighted F1 $\uparrow$ & Cohen's Kappa $\uparrow$ & G-Mean $\uparrow$ & Macro AUC-ROC $\uparrow$\\
\midrule
ContraWR \cite{yang2023self} & 1.6M & 0.4384$\pm$0.0349 & 0.6893$\pm$0.0136 & 0.3912$\pm$0.0237 & -- & --\\
BIOT \cite{yang2024biot} & 3.2M & 0.5281$\pm$0.0225 & 0.7492$\pm$0.0082 & 0.5273$\pm$0.0249 & -- & --\\
LaBraM-Base \cite{jiang2024large} & 5.8M & 0.6173$\pm$0.0193 & 0.7445$\pm$0.0134 & 0.5253$\pm$0.0301 & 0.4370$\pm$0.2027 & 0.8896$\pm$0.0064\\
EEGPT \cite{wangeegpt} & 25M & 0.6082$\pm$0.0134 & \underline{0.8114$\pm$0.0081} & \underline{0.6435$\pm$0.0203} & 0.2929$\pm$0.1864 & \underline{0.9103$\pm$0.0173}\\
CBraMod \cite{wang2024cbramod} & 4.0M & \underline{0.6424$\pm$0.0032} & \textbf{0.8176$\pm$0.0063} & \textbf{0.6581$\pm$0.0127} & \underline{0.5387$\pm$0.0124} & 0.8976$\pm$0.0114\\
\midrule
{EEGDM (Ours, Linear probing)} & 25M &  0.6010$\pm$0.0038  & 0.7006$\pm$0.0053 & 0.4373$\pm$0.0072 & -- & --\\
{EEGDM (Ours, Fine-tuned)} & 25M & \textbf{0.6826$\pm$0.0108} & 0.7329$\pm$0.0191 & 0.5357$\pm$0.0318 & \textbf{0.6443$\pm$0.0406} & \textbf{0.9169$\pm$0.0022}\\
\bottomrule
\end{tabular}
\label{table 3}
\end{table*}

We evaluate the representation learning capabilities of EEGDM through fine-tuning on downstream tasks. \revision{During fine-tuning, the encoder is not frozen and all parameters are jointly optimized with a unified learning rate of $5\times10^{-4}$. No separate or layer-wise learning rates are applied.} Specifically, we predict corresponding labels using a linear predictor based on the outputs of the EEG encoder. Following the common evaluation protocols, we present results under the fine-tuning setting, where the predictor is jointly trained with the EEG encoder (i.e., the pre-trained encoder is not frozen). We employ cross-entropy loss for downstream classification tasks and KL divergence loss for downstream distribution prediction tasks. The hyperparameters for fine-tuning EEGDM on downstream tasks are provided in the Supplementary Material (Table II). We conduct a series of quantitative and qualitative experiments to evaluate the proposed model, demonstrating its strong representation capabilities across diverse downstream datasets. \secondrevision{The source code, preprocessing scripts, and training configurations are publicly available at \url{https://github.com/wangsc18/EEGDM}.}



\subsection{Results and Analysis}
In this section, we show that EEGDM achieves competitive performance across diverse downstream tasks, demonstrating the effectiveness of diffusion-based representation learning for EEG analysis. All experiments are conducted with multiple random seeds, and we report the mean and standard deviation of evaluation metrics. The reported model size corresponds to the downstream networks used for fine-tuning, as the generative module is not involved in downstream tasks. Results are compared against baselines, as shown in Tables \ref{table 3}, \ref{table 4}, \ref{table 5} and \ref{table 6}. Compared to other approaches, EEGDM consistently maintains either the best or second-best performance in key metrics (balanced accuracy or KL divergence).


\revision{Table \ref{table 3} reports the fine-tuning results on TUEV under a cross-distribution transfer setting.
We note that Puah-EEGDM-SSM \cite{puah2025eegdm} is pre-trained directly on TUEV and thus benefits from identical pre-training and downstream data distributions; a controlled same-distribution comparison is provided later in this section.
TUEV exhibits severe class imbalance (up to 80:1), for which sample-weighted metrics can be misleading. We therefore additionally report G-Mean and Macro AUC-ROC as class-balanced metrics.
Our method achieves the highest Balanced Accuracy (68.26\%), G-Mean (64.43\%), and Macro AUC-ROC (91.69\%).
We reproduce LaBraM \cite{jiang2024large}, EEGPT \cite{wangeegpt}, and CBraMod \cite{wang2024cbramod} under identical data splits for fair per-class comparison. While they obtain higher Weighted F1 and Kappa, per-class analysis reveals that this comes at the cost of near-zero recall on the minority class (10--18\%), whereas EEGDM improves the SPSW recall to 50.3\% while maintaining competitive recall on the remaining categories. Detailed per class recall values are provided in the Supplementary Material (Table III).}

\revision{To further isolate the effect of pre-training data distribution, we pre-train EEGDM on the TUEV training set using the identical self-supervised objective and compare it with Puah-EEGDM-SSM under the same-distribution condition.
As shown in Table \ref{tab:tuev_samedist}, our TUEV-pretrained EEGDM achieves a state-of-the-art balanced accuracy of 79.71\%, substantially outperforming Puah-EEGDM-SSM (75.57\%).
This result demonstrates that the performance gap observed in the cross-distribution setting is a byproduct of domain shift between TUEG and TUEV, rather than a limitation of our DiT-based latent diffusion architecture. When evaluated under identical training conditions,} \secondrevision{EEGDM achieves higher balanced accuracy, further supporting the effectiveness of the proposed conditional latent diffusion framework.}

\begin{table}[t]
\centering
\caption{Same-distribution comparison on TUEV. Both methods are pre-trained directly on the TUEV training set.}
\label{tab:tuev_samedist}
\begin{tabular}{lcccc}
\toprule
Method & Balanced Accuracy $\uparrow$ & Weighted F1 $\uparrow$ \\
\midrule
Puah-EEGDM-SSM \cite{puah2025eegdm} & 0.7557$\pm$0.0147 & 0.8688$\pm$0.0066 \\
EEGDM (TUEV-pretrained) & 0.7971$\pm$0.0127 & 0.7690$\pm$0.0362  \\
\bottomrule
\end{tabular}
\end{table}

\begin{table*}[ht]
	\centering
	\caption{Performance comparison on TUAB. The best metric is highlighted in bold, and the second-best metric is underlined.}
	\begin{tabular}{lcccc}
		\toprule
		Methods & Params & Balanced Accuracy $\uparrow$ & AUROC $\uparrow$ & AUC-PR $\uparrow$\\
		\midrule
		ContraWR \cite{yang2023self} & 1.6M & 0.7746$\pm$0.0041 & 0.8456$\pm$0.0074 & 0.8421$\pm$0.0104\\
		BIOT \cite{yang2024biot} & 3.2M & 0.7959$\pm$0.0057 & 0.8815$\pm$0.0043 & 0.8792$\pm$0.0023\\
        
		LaBraM-Base \cite{jiang2024large} & 5.8M & 0.8140$\pm$0.0019 & \underline{0.9022$\pm$0.0009} & 0.8965$\pm$0.0016\\
        EEGPT \cite{wangeegpt} & 25M & 0.7983$\pm$0.0030 & 0.8718$\pm$0.0050 & -- \\
        CBraMod \cite{wang2024cbramod} & 4.0M & \textbf{0.8289$\pm$0.0022} & \textbf{0.9227$\pm$0.0011} & \textbf{0.9258$\pm$0.0008}\\
        Puah-EEGDM-SSM \cite{puah2025eegdm} & 12.8M & 0.8051$\pm$0.0093 & 0.8856$\pm$0.0120 & 0.8981$\pm$0.0092 \\
		\midrule
        {EEGDM (Ours, Linear probing)} & 25M & 0.7813$\pm$0.0170  & 0.8780$\pm$0.0105 & 0.8914$\pm$0.0053 \\
		{EEGDM (Ours, Fine-tuned)} & 25M & \underline{0.8180$\pm$0.0013} & 0.8922$\pm$0.0009 & \underline{0.9085$\pm$0.0002}\\
		\bottomrule
	\end{tabular}
	\label{table 4}
\end{table*}

\begin{table*}[ht]
	\centering
	\setlength{\tabcolsep}{2pt}
	\caption{Performance comparison on DMER. The best result is highlighted in bold, and the second-best result is underlined.}
	\begin{tabular}{lcccccc}
		\toprule
		Methods & Chebyshev $\downarrow$ & Clark $\downarrow$ & Canberra $\downarrow$ & KL $\downarrow$ & Cosine $\uparrow$ & Intersection $\uparrow$ \\
		\midrule
        LaBraM-Base \cite{jiang2024large} & \underline{0.0925$\pm$0.0064} & \textbf{0.2924$\pm$0.0048} & 1.7476$\pm$0.0287 & \underline{0.0995$\pm$0.0038} & \underline{0.9094$\pm$0.0037} & \underline{0.8198$\pm$0.0034} \\
        CBraMod \cite{wang2024cbramod} & 0.0945$\pm$0.0052 & 0.3024$\pm$0.0286 & 1.8320$\pm$0.1984 & 0.1077$\pm$0.0148 & 0.9026$\pm$0.0116 & 0.8104$\pm$0.0184\\
        EEGPT \cite{wangeegpt} & 0.1016$\pm$0.0020 & 0.3009$\pm$0.0055 & \underline{1.7264$\pm$0.0204} & 0.1055$\pm$0.0045 & 0.9005$\pm$0.0042 & 0.8172$\pm$0.0026\\
        Puah-EEGDM-SSM \cite{puah2025eegdm} & 0.0940$\pm$0.0028 & 0.3012$\pm$0.0057 & 1.8275$\pm$0.0687 & 0.1032$\pm$0.0015 & 0.9072$\pm$0.0006 & 0.8132$\pm$0.0040 \\
		\midrule
        {EEGDM (Ours, Linear probing)} & 0.0942$\pm$0.0009  & 0.3091$\pm$0.0109 & 1.9099$\pm$0.0984 & 0.1102$\pm$0.0066 & 0.9023$\pm$0.0036 & 0.8058$\pm$0.0070 \\
		{EEGDM (Ours, Fine-tuned)} & \textbf{0.0836$\pm$0.0078} & \underline{0.2984$\pm$0.0053} & \textbf{1.6112$\pm$0.0108} & \textbf{0.0978$\pm$0.0063} & \textbf{0.9176$\pm$0.0083} & \textbf{0.8258$\pm$0.0047} \\
		\bottomrule
	\end{tabular}
	\label{table 5}
\end{table*}

\begin{table*}[ht]
\centering
	\caption{Performance comparison on SHU-MI, CL-Drive and Seed-VII datasets. The best result is highlighted in bold, and the second-best result is underlined. Results for Puah-EEGDM-SSM on SHU-MI are omitted due to performance near chance levels.}
	\begin{tabular}{lcccc}
		\toprule
		Datasets & Methods & Balanced Accuracy $\uparrow$ & Weighted F1 / AUROC $\uparrow$ & Cohen’s Kappa / AUC-PR $\uparrow$ \\
		\midrule
		\multirow{6}{*}{SHU-MI} 
		& LaBraM-Base \cite{jiang2024large} & 0.6166$\pm$0.0192 & 0.6604$\pm$0.0091 & 0.6761$\pm$0.0083 \\
		& CBraMod \cite{wang2024cbramod} & \textbf{0.6370$\pm$0.0151} & \underline{0.6988$\pm$0.0068} & \textbf{0.7139$\pm$0.0088} \\
        & EEGPT \cite{wangeegpt} & 0.5565$\pm$0.0345 & 0.5908$\pm$0.0502 & 0.5996$\pm$0.0491 \\
        \cmidrule(lr){2-5}
        & {EEGDM (Ours, Linear probing)} & 0.5319$\pm$0.0061 & 0.5463$\pm$0.0049 & 0.5400$\pm$0.0093 \\
		& {EEGDM (Ours, Fine-tuned)} & \underline{0.6324$\pm$0.0052} & \textbf{0.6996$\pm$0.0035} & \underline{0.7070$\pm$0.0061} \\
		\midrule
        
		\multirow{6}{*}{CL-Drive} 	
		& LaBraM-Base \cite{jiang2024large}& 0.4096$\pm$0.0942 & 0.4438$\pm$0.0972 & 0.0745$\pm$0.1087 \\
		& CBraMod \cite{wang2024cbramod} & \underline{0.5346$\pm$0.0856} & 0.4896$\pm$0.1237 & 0.2322$\pm$0.1014 \\
        & EEGPT \cite{wangeegpt} & 0.4674$\pm$0.0562 & \underline{0.5038$\pm$0.0640} & 0.1641$\pm$0.0894 \\
        & Puah-EEGDM-SSM \cite{puah2025eegdm} & 0.5092$\pm$0.0457 & \textbf{0.5513$\pm$0.0408} & \underline{0.2570$\pm$0.0676} \\
        \cmidrule(lr){2-5}
        & {EEGDM (Ours, Linear probing)} & 0.5295$\pm$0.0272  & 0.4585$\pm$0.0545 & 0.2275$\pm$0.0394 \\
		& {EEGDM (Ours, Fine-tuned)} & \textbf{0.5357$\pm$0.0557} & 0.5019$\pm$0.0725 & \textbf{0.2944$\pm$0.0841} \\
        \midrule
        
        \multirow{6}{*}{Seed-VII} 
		& LaBraM-Base \cite{jiang2024large}& \underline{0.2281$\pm$0.0077} & \textbf{0.2198$\pm$0.0113} & \textbf{0.1003$\pm$0.0090} \\
		& CBraMod \cite{wang2024cbramod} & 0.2027$\pm$0.0045 & 0.1824$\pm$0.0034 & 0.0731$\pm$0.0041 \\
        & EEGPT \cite{wangeegpt} & 0.2125$\pm$0.0112 & 0.1989$\pm$0.0264 & 0.0849$\pm$0.0159 \\
        & Puah-EEGDM-SSM \cite{puah2025eegdm} & 0.1810$\pm$0.0007 & 0.1386$\pm$0.0050 & 0.0516$\pm$0.0012 \\
        \cmidrule(lr){2-5}
        & {EEGDM (Ours, Linear probing)} & 0.2151$\pm$0.0221  & 0.1875$\pm$0.0249 & 0.0818$\pm$0.0229 \\
		& {EEGDM (Ours, Fine-tuned)} & \textbf{0.2300$\pm$0.0021} & \underline{0.2115$\pm$0.0087} & \underline{0.0943$\pm$0.0395} \\
		\bottomrule
	\end{tabular}
	\label{table 6}
\end{table*}

\begin{table}[t]
\centering
\caption{Training paradigm comparison on TUEV. All use the same EEG-ViT-Large encoder architecture (25M).}
\label{tab:paradigm}
\begin{tabular}{lcc}
\toprule
Training Paradigm & Balanced Accuracy $\uparrow$ & Weighted F1 $\uparrow$ \\
\midrule
From scratch & 0.5825$\pm$0.0116 & 0.7072$\pm$0.0065 \\
Linear probing & 0.6010$\pm$0.0038 & 0.7006$\pm$0.0053 \\
Fine-tuned & 0.6826$\pm$0.0108 & 0.7329$\pm$0.0191 \\
\bottomrule
\end{tabular}
\end{table}

\begin{table}[!htbp]
\centering
\setlength{\tabcolsep}{2pt}
	\caption{Performance comparison of two data augmentation strategies on the CL-Drive dataset.}
	\begin{tabular}{cccc}
		\toprule
		Strategy & Balanced Accuracy ↑ & Weighted F1 ↑ & Cohen’s Kappa ↑ \\
		\midrule
        None & 0.5357$\pm$0.0557 & 0.5019$\pm$0.0725 & 0.2944$\pm$0.0841 \\
		Proportional & 0.5896$\pm$0.0349 & 0.5579$\pm$0.0304 & 0.3169$\pm$0.0448 \\
		Balanced & 0.5585$\pm$0.0278 & 0.6047$\pm$0.0267 & 0.3319$\pm$0.0430 \\
		\bottomrule
	\end{tabular}
	\label{table 8}
\end{table}

Table \ref{table 4} shows that EEGDM achieves a balanced accuracy of 81.80\% and an AUC-PR of 90.85\% on the TUAB dataset. Securing the second-best performance on those metrics, EEGDM overall demonstrates comparable performance with a competitive margin. The consistent performance across both balanced accuracy and AUC-PR metrics suggests that EEGDM learns robust representations that generalize well for binary abnormal detection tasks, maintaining high discriminative power even in challenging clinical scenarios.

Table \ref{table 5} presents the results for the DMER dataset, which uses distributional metrics. EEGDM achieves best performance across various divergence and similarity metrics, with the exception of the Clark metric, where it ranks second. The overall high performance on DMER, a task requiring the prediction of a continuous emotion distribution rather than a discrete class, strongly suggests that the diffusion-based pre-training learns rich and fine-grained features that preserve the complex, subtle, and continuous emotional states embedded in the EEG signals.

\begin{table*}[!htbp]
	\centering
	\setlength{\tabcolsep}{4.5pt}
	\caption{Ablation results of PCA and EEG augmentation on TUEV. All metrics are reported as mean $\pm$ standard deviation over three random seeds.}
	\begin{tabular}{cccccc}
		\toprule
		Variants & PCA & EEG Augmentation & Balanced Accuracy $\uparrow$ & G-Mean $\uparrow$ & Macro AUC-ROC $\uparrow$ \\
		\midrule
		A & \ding{55} & \ding{55} & $0.6054\pm0.0083$ & $0.3391\pm0.2155$ & $0.8750\pm0.0059$ \\
		B & \ding{51} & \ding{55} & $0.5805\pm0.0060$ & $0.4578\pm0.0496$ & $0.8592\pm0.0074$ \\
		C & \ding{55} & \ding{51} & $0.5917\pm0.0020$ & $0.4544\pm0.0717$ & $0.8547\pm0.0187$\\
		D & \ding{51} & \ding{51} & $0.6826\pm0.0108$ & $0.6443\pm0.0406$ & $0.9169\pm0.0022$ \\
		\bottomrule
	\end{tabular}
	\label{table 7}
\end{table*}

Table \ref{table 6} further validates the generalizability and robustness of EEGDM across diverse BCI tasks. Specifically, on the SHU-MI dataset, EEGDM ranks second in balanced accuracy (63.24\%) and AUC-PR (70.70\%), while achieving the best performance in AUROC (69.96\%), demonstrating competitive performance compared with CBraMod.
\revision{Notably, the results for Puah-EEGDM-SSM \cite{puah2025eegdm} on this specific benchmark are omitted from the table, as its performance fluctuated near chance levels (e.g., accuracy $\approx 50\%$). This discrepancy suggests that while domain-specific pre-training (as used in \cite{puah2025eegdm}) may excel in aligned tasks, our architecture, pre-trained on a broader corpus, exhibits superior cross-task adaptability, particularly in capturing the subtle spatial-temporal patterns required for motor imagery.} 
For the CL-Drive dataset, EEGDM shows the best balanced accuracy and the best Cohen’s Kappa, demonstrating superior cognitive load feature extraction capabilities. On the Seed-VII dataset, EEGDM continues its strong performance with the best performance in balanced accuracy and ranks second in weighted F1 and Cohen’s Kappa, further validating its effectiveness in emotion-related tasks.

\revision{To isolate the contribution of diffusion-based pretraining, we compare three paradigms on TUEV using the same EEG-ViT-Large backbone (25M): training from random initialization (from scratch), linear probing with a frozen encoder, and full fine-tuning. As shown in Table \ref{tab:paradigm}, the from-scratch model achieves only 58.25\% balanced accuracy — over 9 points below the fine-tuned pretrained encoder. This gap confirms that the performance advantage of EEGDM derives from pretraining rather than model capacity.}

\revision{A unique capability of diffusion-based pretraining is the ability to synthesize new EEG samples. We validate this on CL-Drive, a small-scale driving behavior dataset with class imbalance. Synthetic EEG samples are generated by conditioning the diffusion model on encoder representations of real data, then merged with the original training set. We evaluate two augmentation strategies: proportional, which preserves the original class distribution, and balanced, which generates an equal number of samples per class. As shown in Table \ref{table 8}, the proportional strategy yields the best balanced accuracy (58.96\%), improving over the unaugmented baseline (53.57\%) by 5.4 points. The balanced strategy achieves the highest Weighted F1 (60.47\%) and Cohen's Kappa (33.19\%), indicating a different trade-off between overall classification performance and balanced accuracy. Both strategies outperform the baseline, demonstrating the practical utility of EEGDM as a generative data augmentation framework.}

These comprehensive results across diverse datasets validate the effectiveness of using EEG signal generation as a self-supervised pre-training objective, demonstrating that the combination of encoder and diffusion denoiser, enhanced by PCA and channel augmentation, successfully learns high-quality EEG representations for improved downstream task performance.

\subsection{Ablation Study}
\label{Ablation}

\secondrevision{We conduct ablation studies to examine the effects of PCA and EEG augmentation using the four variants shown in Table~\ref{table 7}. All downstream results on TUEV are reported as the mean and standard deviation over three random seeds. When applied independently, the PCA-only and augmentation-only variants yield slightly lower balanced accuracy and Macro AUC-ROC than the baseline, suggesting no improvement in overall discriminative performance on these metrics.}

\secondrevision{Interestingly, the G-Mean results reveal a different aspect of their effects. The baseline obtains a G-Mean of $33.91\pm21.55\%$, whereas PCA only and augmentation only achieve $45.78\pm4.96\%$ and $45.44\pm7.17\%$, respectively, indicating substantially improved class-wise performance stability. Notably, in one of the three random seeds, the baseline without PCA or augmentation yields a G-Mean of approximately $0.03$. Since G-Mean is the geometric mean of the per-class recalls, this indicates that at least one EEG event class has a recall no greater than approximately $3\%$, despite the relatively competitive average balanced accuracy of the baseline. When PCA and EEG augmentation are jointly applied, the full model achieves the strongest overall performance, with a balanced accuracy of $68.26\%$, a G-Mean of $64.43\%$, and a Macro AUC-ROC of $91.69\%$. 
A possible explanation for these observations lies in the different roles of the two components within the conditional diffusion framework. PCA operates on the denoising target by projecting EEG patches into a compact latent space, whereas EEG augmentation acts on the conditioning branch by encouraging the encoder to preserve information shared across perturbed views. When jointly used, the information retained by the conditional representation may become better aligned with the dominant structure preserved in the PCA target space, which is consistent with the improved class-balanced performance. Additional representation-space evidence based on t-SNE visualization and FDR analysis is provided in the Supplementary Material.}

\begin{figure}[!htb]
	\centering
	\subfigure[TUEV]{
		\label{Figure 3(a)} 
		\includegraphics[width=1.6in]{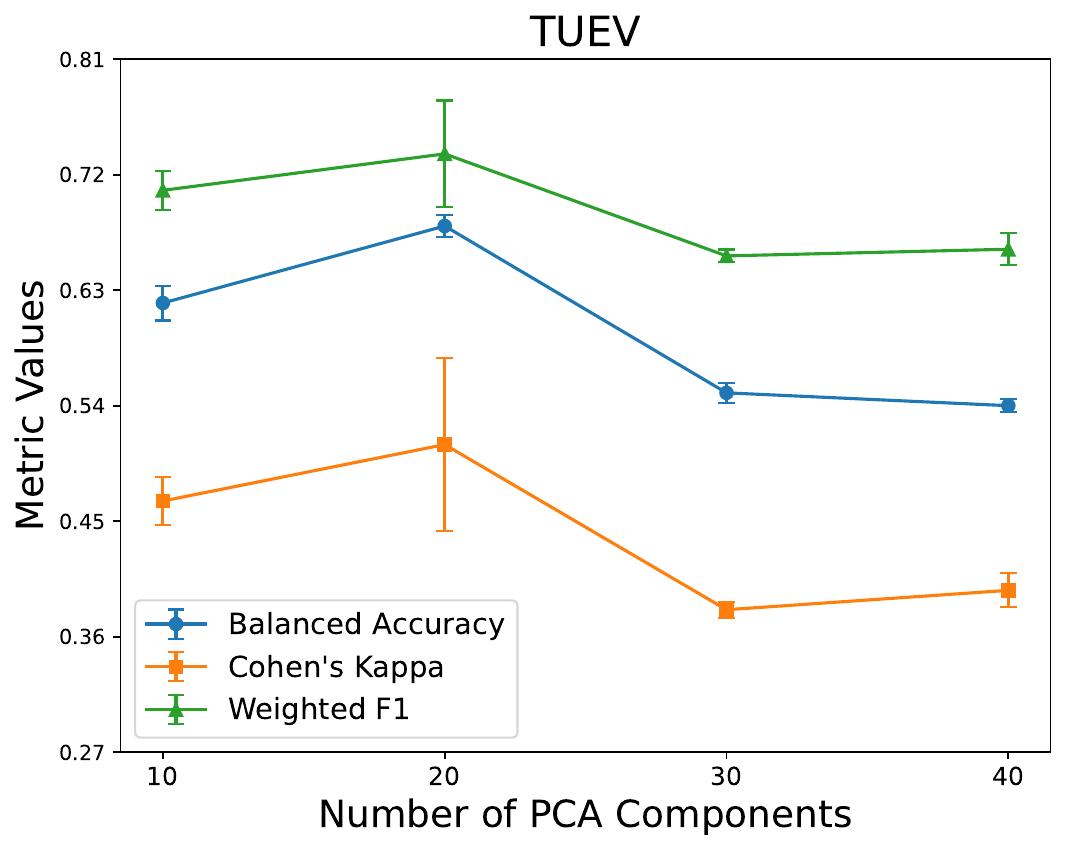}}
	\hspace{0.00in}
	\subfigure[TUAB]{
		\label{Figure 3(b)} 
		\includegraphics[width=1.6in]{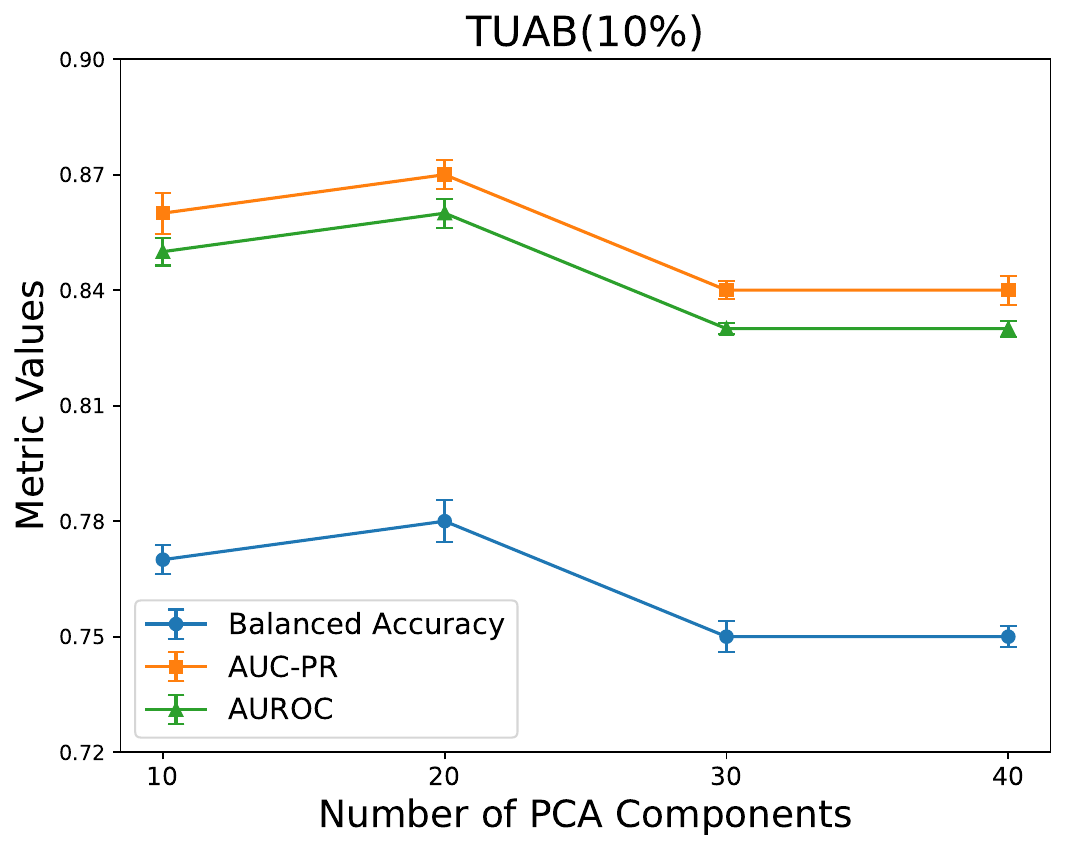}}
	\caption{Performance comparison with different numbers of PCA components on TUEV and TUAB datasets.}
	\label{Figure 3}  
\end{figure}

Beyond the effects of PCA and EEG augmentation, we further investigate the influence of PCA dimensionality.
Fig. \ref{Figure 3} presents results on TUEV and TUAB with varying numbers of PCA components, which reveal a nonlinear relationship between the number of PCA components and downstream task performance. In particular, insufficient number of components may result in large information loss, whereas an excessive number introduces noise contamination, both of which degrade model performance. To balance these trade-offs, we empirically select an intermediate value of 20 components, which yields optimal performance in our experiments.

\section{Conclusion}
\label{section5}
We present EEGDM, a self-supervised framework that leverages diffusion-based EEG generation as a representation learning objective. \secondrevision{By using the EEG encoder representation as conditional information for the diffusion model, the encoder and diffusion denoiser are jointly optimized through the generative objective to learn inherent EEG representations.} 
EEGDM achieves competitive performance across six diverse EEG analysis tasks spanning clinical diagnostics, brain-computer interfaces, and emotion recognition, \secondrevision{with particularly strong class-balanced performance on the severely imbalanced TUEV dataset. Ablation and representation-space analyses further show that the joint use of PCA-based temporal compression and EEG augmentation improves representation discriminability and class-balanced robustness. }
Beyond representation learning, EEGDM's diffusion module enables synthetic EEG generation for downstream data augmentation, a capability unavailable to masked-reconstruction models. These results demonstrate that diffusion-based generation offers a viable and complementary paradigm for EEG foundation models.

\bibliographystyle{IEEEtran}  
\bibliography{mybibfile}

\clearpage
\onecolumn
\appendices
\section*{Supplementary Material}

\subsection{Dataset Details for Downstream Tasks}
This section provides a detailed description of the downstream datasets employed to evaluate the performance of our proposed model.

(1) \textbf{TUEV} (event type classification) \cite{harati2015improved}: An EEG dataset containing six event categories: spike and sharp wave (SPSW), generalized periodic epileptiform discharges (GPED), periodic lateralized epileptiform discharges (PLED), eye movement (EYEM), artifact (ARTF), and background (BCKG). The EEG signals are recorded with 23 channels at a sampling rate of 256 Hz.

(2) \textbf{TUAB} (abnormal detection) \cite{lopez2015automated}: A corpus of EEG recordings annotated as either normal or abnormal. Similar to TUEV, the EEG signals in TUAB are recorded with 23 channels at a sampling rate of 256 Hz.

(3) \textbf{SHU-MI} (motor imagery classification) \cite{ma2022large}: An EEG motor imagery dataset designed for decoding imagined movements. It features 32 EEG channels sampled at 250 Hz, including two types of motor imagery tasks: left hand and right hand.

(4) \textbf{CL-Drive} (cognitive load assessment) \cite{angkan2024multimodal}: A multimodal dataset designed for cognitive load assessment, which includes EEG signals. The EEG data are recorded using four channels at a sampling frequency of 256 Hz. This study explores the ternary classification of cognitive load into high, medium, and low levels.

(5) \textbf{Seed-VII} (emotion recognition) \cite{jiang2024seed}: A large-scale multimodal emotion dataset. Seed-VII includes 62-channel EEG and eye-movement data from 20 healthy subjects across seven emotion categories (disgust, fear, sad, neutral, happy, anger, surprise). The preprocessed EEG signals were downsampled to 200 Hz.

(6) \textbf{DMER} (emotion distribution prediction) \cite{yang2024multimodal}: A multimodal emotion recognition corpus that includes EEG signals along with other modalities (GSR, PPG, and frontal facial videos). In each trial, participants watched video clips and subsequently completed a self-assessment using a ten-item version of the Positive Affect and Negative Affect Schedule (PANAS) \cite{mackinnon1999short}, reporting the intensity of ten emotional adjectives. Each emotion rating ranges from 1 (not at all) to 5 (extremely), and the resulting scores are transformed into a 10-dimensional emotion distribution. The EEG recordings comprise 18 channels sampled at 300 Hz.

\begin{table}[h]
	\centering
	\caption{Hyperparameters for EEGDM pre-training.}
	\begin{tabular}{ccc}
		\toprule
		& Hyperparameter & Value \\
		\hline
		\multirow{11}{*}{Optimization} & Batch size & 32 \\
		& Peak learning rate & 5e-4 \\
		& Minimal learning rate & 5e-5 \\
		& Learning rate scheduler & OneCycleLR \\
		& Weight decay & 0.05 \\
		& EMA decay rate & 0.9999 \\
		& Gradient clipping norm & 1 \\
		& Optimizer & AdamW \\
		& ${\beta}_{adam}$ & (0.9, 0.999) \\
        & $\lambda_{\mathrm{vlb}}$ & 1 \\
		\hline
		\multirow{4}{*}{Diffusion Module} & Layers N & 8 \\ 
		& Hidden size & 512 \\
		& Heads & 8 \\
		& Normalization type & LayerNorm \\
        & Diffusion training steps & 1000 \\
		\hline
		\multirow{4}{*}{EEG Encoder} & Layers & 8 \\
		& Hidden size & 512 \\
		& MLP size & 2048 \\
		& Heads & 8 \\
		\bottomrule
	\end{tabular}%
	\label{tables1}%
\end{table}%

\begin{table}[h]
  \centering
  \caption{Hyperparameters for downstream fine-tuning and linear probing. ($\star$) Depends on the dataset.}
 \label{Hyperparameters for downstream fine-tuning and linear probing}
    \begin{tabular}{lcc}
   \toprule
   Hyperparameter & Fine-tuning & Linear Probing \\ 
   \hline
   Encoder status & Unfrozen & Frozen \\
   Learning rate$^\star$ & 5e-4 & 2e-4, 5e-4, 2e-3 \\
   Total epochs$^\star$ & 30, 150 & 30, 50, 100 \\
   Weight decay$^\star$ & 0.05 & 0.05, 0.02, 0.01 \\
   Batch size$^\star$ & 64 & 32, 64 \\
   Optimizer & AdamW & AdamW \\ 
   Learning rate scheduler$^\star$ & OneCycleLR & OneCycleLR, ReduceLROnPlateau, CosineAnnealingLR \\
   ${\beta}_{adam}$ & (0.9, 0.999) & (0.9, 0.999) \\
   Warmup epochs$^\star$ & 6, 30 & 3, 10 \\
\bottomrule
\label{tables2}
\end{tabular}
\end{table}

\begin{table}[h]
\centering
\caption{Per-class recall on TUEV. Values are means over three random seeds.}
\label{tab:perclass_recall}
\begin{tabular}{lcccccc}
\toprule
Method & SPSW (0) & GPED (1) & PLED (2) & EYEM (3) & ARTF (4) & BCKG (5) \\
\midrule
EEGPT & 0.149 & 0.803 & 0.476 & 0.887 & 0.614 & 0.876 \\
LaBraM & 0.180 & 0.751 & 0.458 & 0.866 & 0.590 & 0.863 \\
CBraMod & 0.123 & 0.838 & 0.479 & 0.905 & 0.625 & 0.885 \\
EEGDM (Ours) & 0.503 & 0.792 & 0.491 & 0.859 & 0.581 & 0.856 \\
\bottomrule
\label{tables3}
\end{tabular}
\end{table}

\subsection{Training Hyperparameters}

Tables~\ref{tables1} and \ref{tables2} list the hyperparameters used for pre-training and downstream fine-tuning, respectively.

\begin{figure}[htb]
\centering
\includegraphics[width=\textwidth]{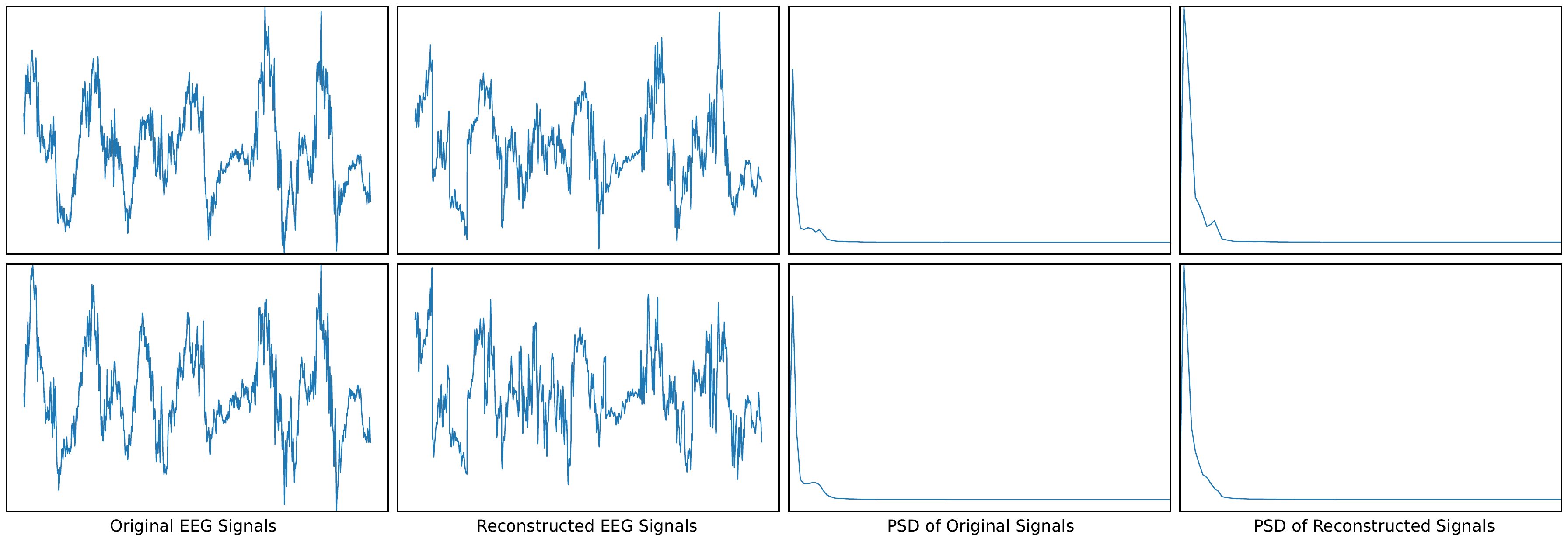}
\caption{Original and reconstructed EEG signals with PSD comparison.}
\label{fig:S1}
\end{figure}

\begin{figure}[htb]
    \centering
    \includegraphics[width=3.5in]{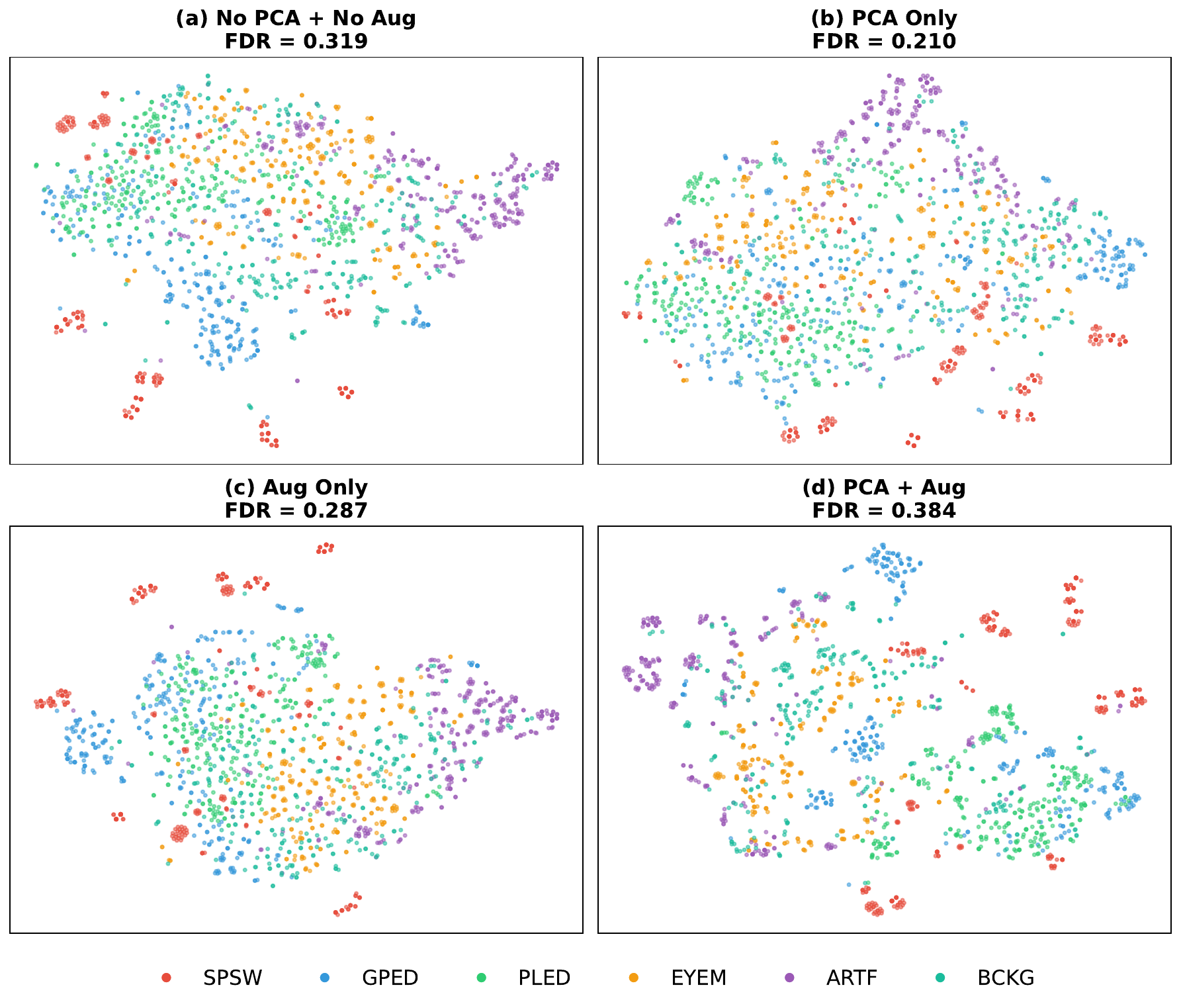}
    \caption{t-SNE visualizations of representations extracted by the four pre-trained encoder variants on the TUEV test set without downstream fine-tuning: (a) no PCA + no augmentation, (b) PCA only, (c) augmentation only, and (d) PCA + augmentation. We randomly sample 300 test samples from each class and use the identical samples for all four variants. The reported FDR values are computed directly from the corresponding 512-dimensional encoder representations rather than from the two-dimensional t-SNE embeddings.}
    \label{Figure 4}
\end{figure}

\subsection{Per-Class Performance on TUEV}

Table \ref{tables3} reports the per-class recall of all compared methods. EEGDM achieves over 50\% recall on the rarest class (SPSW, class ratio up to 80:1), while masked-reconstruction baselines attain only 10--18\%.

\subsection{EEG Signal Reconstruction}

Fig.~\ref{fig:S1} shows reconstructed EEG signals from the TUEG pre-training set alongside their power spectral density (PSD). Both the temporal waveform and dominant neural rhythms closely match the ground truth.

\subsection{Representation Visualization}

{To further complement the ablation analysis in the main manuscript, we examine whether the performance differences among the four variants are already reflected in the representations learned during self-supervised pre-training. We analyze the four encoder variants on TUEV without any downstream fine-tuning. Specifically, we randomly sample 300 test samples from each event class and use the identical samples for all four variants. The representations are directly extracted from the pre-trained encoders and visualized using t-SNE~\cite{van2008visualizing}, as shown in Fig.~\ref{Figure 4}.}

{In addition to the qualitative visualization, we report the Fisher discriminant ratio (FDR) computed directly in the original 512-dimensional encoder representation space. FDR measures class separability by comparing inter-class variation with intra-class variation, with a larger value indicating stronger class-discriminative structure. The no-PCA/no-augmentation, PCA-only, and augmentation-only variants obtain FDR values of $0.319$, $0.210$, and $0.287$, respectively, whereas the full PCA + augmentation model achieves the highest FDR of $0.384$.}

{Consistent with the FDR results, the full model exhibits a more structured class-dependent organization in the t-SNE visualization, although substantial overlap remains among the heterogeneous EEG event classes. Since the representations are extracted directly from the pre-trained encoders without downstream fine-tuning, these results suggest that the improved class-discriminative structure is already learned during self-supervised pre-training.}

\end{document}